\documentclass{opt2020_style/opt2020}
\usepackage{bm}
\usepackage{wrapfig}
\usepackage{cleveref}
\newcommand{\pdfAuthor}       {Mark Tuddenham}
\newcommand{\pdfTitle}        {Quasi-Newton's method in the class gradient defined high-curvature subspace}
\newcommand{\pdfSubject}      {Optimisation of Deep Learning}
\newcommand{\pdfKeywords}     {Deep Learning, Machine Learning, Optimisation, Quasi-Newtons, high-curvature}
\newcommand{\pdfCreator}      {\pdfAuthor}
\newcommand{\pdfProducer}     {\pdfCreator}
\hypersetup{
  pdfinfo={
    Title={\pdfTitle{}},
    Author={\pdfAuthor{}},
    Subject={\pdfSubject{}},
    Keywords={\pdfKeywords{}},
    Creator={\pdfCreator{}},
    Producer={\pdfProducer{}},
    }
}
\title{\pdfTitle{}}
\optauthor{\Name{Mark Tuddenham} \Email{mark.tuddenham@soton.ac.uk}\\
  \Name{Adam Prügel-Bennett} \Email{apb@ecs.soton.ac.uk}\\
  \Name{Jonathan Hare} \Email{jsh2@ecs.soton.ac.uk}\\
  \addr University of Southampton,  Southampton,  SO17 1BJ,  United Kingdom}
\begin{document}
\maketitle

\newcommand{\logg}[1]{\log\!\left(#1 \right)}
\newcommand{\av}[2][]{\mathbb{E}_{#1\!}\left[ #2 \right]}
\newcommand{\Prob}[2][]{\mathbb{P}_{#1\!}\left[ #2 \right]}
\newcommand{\grad}{\bm{\nabla}}
\newcommand{\tr}{\mathsf{T}}
\newcommand{\mat}[1]{\boldsymbol{\mathsf{#1}}}
\newcommand{\e}[1]{{\rm e}^{#1}} 
\newcommand{\ie}{\textit{i.e.}}
\newcommand{\eg}{\textit{e.g.}}
\newcommand{\inc}{\textit{inc.}}
\newcommand{\etc}{\textit{etc.}}
\newcommand{\wrt}{\textit{w.r.t.}}
\newcommand{\Hlr}{H^{\text{\small Low Rank}}}
\begin{abstract}
    Classification problems using deep learning have been shown to have a high-curvature subspace in the loss landscape equal in dimension to the number of classes.
    Moreover, this subspace corresponds to the subspace spanned by the logit gradients for each class.
    An obvious strategy to speed up optimisation would be to use Newton's method in the high-curvature subspace and stochastic gradient descent in the co-space.
    We show that a naive implementation actually slows down convergence and we speculate why this might be.
\end{abstract}

\section{Introduction}

Optimisation is often not a prime consideration for practitioners training deep neural networks. This is because there are many options where the large majority of choices optimise well but getting the best optimisation needs a seemingly arbitrary choice of (almost always first-order) update method, learning rate, decay schedule, momentum, \etc{}

The ideal combination of these options has mainly been brute-force searched for each individual problem: see the research done on speed training \eg{} DAWNBench~\cite{Coleman2017}.
This shows it is difficult to choose the ideal combination when faced with a new problem or a new network architecture.
Moreover, as this ideal choice is often dependant on the curvature of the loss landscape, second-order methods, which are designed to capture the local curvature, should be more powerful and robust than first-order methods.

\subsection{Second-order methods}
Directions of high curvature limit the step size that can be taken in gradient descent.
To see this, consider a quadratic minimum defined by a Hessian, $H$,
\begin{equation}
    f(\theta) = \frac{1}{2} {(\theta-\theta^*)}^\tr H\, (\theta-\theta^*).
\end{equation}
We can write the Hessian matrix as its eigen-decomposition
\begin{align}
    H = \sum_i \lambda_i v_i\,v^\tr_i
\end{align}
where $\lambda_i$ is a measure of the curvature in the direction $v_i$.
For $\theta^*$ to be a minimum we require $\lambda_i>0$ for all $i$.
If we perform gradient descent, $\theta^{(t+1)} = \theta^{(t)} - \eta\, \grad f(\theta^{(t)}) = \theta^{(t)} - \eta\, H\,(\theta^{(t)}-\theta^*)$, then
\begin{align}
    \theta^{(t+1)} - \theta^*
     & = \left( \mat{I}-\eta\,H \right) \left( \theta^{(t)}-\theta^* \right)
    = {\left( \mat{I}-\eta\,H \right)}^t \left( \theta^{(1)}-\theta^* \right)                \\
     & = \sum_i v_i \, {(1-\eta\,\lambda_i)}^t v_i^\tr \left( \theta^{(1)}-\theta^* \right).
\end{align}
Thus gradient descent will diverge exponentially fast unless $\eta\leq 2/\lambda_{\text{\footnotesize max}}$.

Second-order methods are a form of preconditioned gradient methods using second derivatives; these methods work well for ill-conditioned problems as the Hessian ``automatically normalize[s] the ill-conditioned problem by stretching and contracting''\cite{Yao2020} the directions corresponding to its eigenvectors.
However, the difficulty in calculating accurate second-order information holds this class of methods back from improving general deep learning optimisation.
If there are only a small number of directions in weight space with high curvature, then we only need to precondition the gradient in those directions.

\subsection{A second-order method}

\citet{GurAri2018} show that gradient descent is mostly contained in a small subspace, so although the problem is high-dimensional, the optimisation is limited by a low-dimensional high-curvature subspace.
That subspace, spanned by the logit gradients, intersects with the top-$C$ eigenvectors of the Hessian~\cite{Papyan2019, Ghorbani2019}.
If there are a small number of identifiable directions in weight space with high curvature, then we can significantly speed up optimisation by performing Newton's method in this high-curvature subspace and a standard first-order method in the low-curvature dual subspace.

\subsection{Setting}
We consider a classification problem with data, $\mathcal{D} = \left\{ (x^{\mu},y^{\mu})\, \big|\, \mu\in[1,N]\right\}$ where \(y^{\mu}\) is a $C$-dimensional one hot vector, as in \citet{Fort2019}.

We will look at some deep network with parameters, $\theta$, that calculates logits, $z_k^\mu$, which gives class probabilities, $p_k^\mu$, via a softmax.
We will train this network with a cross-entropy loss, $\mathcal{L}^\mu(\theta)$ for an input $(x^{\mu},y^{\mu}) \in \mathcal{D}$.
Then, for any single parameter $\theta_\alpha$ we have its gradient, $g_{\alpha}^{\mathcal{B}}$,\footnote{We derive this as the negative of what \citet{Fort2019} state, however, this disparity only affects the parts of the Hessian that we ignore.} for some minibatch
\begin{equation}
    g_{\alpha}^{\mathcal{B}} = \frac{\partial \mathcal{L}(\theta)}{\partial \theta_\alpha}
    = \frac{1}{|\mathcal{B}|} \sum_{\mu \in \mathcal{B}}
    \sum_{k=1}^C  (p_k^{\mu} - y_k^{\mu})\,
    \frac{\partial z_k^{\mu}}{\partial \theta_\alpha},
\end{equation}
and so the Hessian has components
\begin{align}
    H^{\mathcal{B}}_{\alpha\beta} =
    \frac{\partial^2 \mathcal{L}(\theta)}{\partial \theta_\alpha\partial \theta_\beta}
     & = \frac{1}{|\mathcal{B}|} \sum_{\mu \in \mathcal{B}}
    \sum_{k=1}^C \sum_{\ell=1}^C
    p_k^\mu \left(  \delta_{k\ell} - \,p_\ell^\mu \right)
    \frac{\partial z_k^{\mu}}{\partial \theta_\alpha}
    \frac{\partial z_\ell^{\mu}}{\partial \theta_\beta} \nonumber \\
     & \quad +\frac{1}{|\mathcal{B}|} \sum_{\mu \in \mathcal{B}}
    \sum_{k=1}^C   (p_k^{\mu} - y_k^{\mu})\,
    \frac{\partial^2 z_k^{\mu}}{\partial \theta_\alpha\partial \theta_\beta}
    \label{eq:Hessian}
\end{align}
\citet{Fort2019} claim that the logits are almost a linear function of the weights \textit{i.e.}
\begin{equation}
    \frac{\partial^2 z_k^{\mu}}{\partial \theta_\alpha \partial \theta_\beta} \approx 0
\end{equation}
and so the second term of the Hessian, \cref{eq:Hessian}, can be ignored~\cite{BenaychGeorges2011}.
This term will be exactly 0 for a piecewise linear network \eg{} with ReLUs.

\subsection{Class gradients}
\citet{Fort2019} define the class gradient as
\begin{equation}
    c_k = \frac{1}{| \mathcal{D}_k|}\sum_{\mu\in \mathcal{D}_k} \grad z_k^{\mu}
\end{equation}
where \(\mathcal{D}_k\) is the set of training examples belonging to class \(k\).
The logit gradients can be decomposed as
\begin{equation}
    \grad z_k^{\mu} = y^{\mu}_k\,c_k + \epsilon_k^{\mu}.
    \label{eq:logits}
\end{equation}
Substituting the logit gradient decomposition, \cref{eq:logits}, into the first term of the Hessian, \cref{eq:Hessian}, we get
\begin{equation}
    H \approx
    \sum_{k=1}^C \left( \frac{1}{|\mathcal{B}|}
    \sum_{\mu \in \mathcal{B}} y_k^\mu \,
    p_k^{\mu}(1 - p_k^{\mu})   \right)       c_k\, c_k^\tr
    + \frac{1}{|\mathcal{B}|} \sum_{\mu \in \mathcal{B}}
    \sum_{k=1}^C \sum_{\ell=1}^C
    p_k^\mu \left( \delta_{k\ell} - p_\ell^\mu \right)
    \epsilon^\mu_k\, \epsilon^\mu_\ell
    \label{eq:hessian_approx}
\end{equation}
where, following \citet{Fort2019}, we have ignored the second term in the Hessian and the cross term --- as long as a term is sufficiently small the eigensystem of the Hessian remains sufficiently unchanged~\cite{BenaychGeorges2011}.
Thus a low-rank approximation of the Hessian is the first term in \cref{eq:hessian_approx}, now denoted $\Hlr$.

\section{Quasi-Newton's method}
Assuming that the class gradients are orthogonal we have
\begin{equation}
    \Hlr = \sum_{k=1}^C \lambda_k v_k \, v_k^\tr,
\end{equation}
where $v_k = c_k/|c_k|$ and
\begin{equation}
    \lambda_k = \frac{1}{|\mathcal{B}|}
    \sum_{\mu \in \mathcal{B}} y_k^\mu \,
    p_k^\mu \,(1-p_k^\mu)  \, |c_k|^2.
\end{equation}
As this is a sum of non-negative elements it is positive definite which is a necessary condition for Newton's method to move downhill. As $\Hlr$ is singular it has no inverse, however, we can define a generalised inverse
\begin{equation}
    {(\Hlr)}^\dagger = \sum_{k=1}^C \frac{1}{\lambda_k} v_k \, v_k^\tr.
\end{equation}
We can use this generalised inverse to perform Newton's method in the high-curvature subspace, $\theta' \leftarrow \theta - {(\Hlr)}^\dagger g^\mathcal{B}$.
To perform stochastic gradient descent in the orthogonal subspace we can use the projection operator
\begin{equation}
    \mathbf{P} = \mat{I} - \sum_{k=1}^C v_k\,v_k^\tr
\end{equation}
applied to the gradient $\theta' \leftarrow \theta - \eta \mathbf{P} g^\mathcal{B}$, where $\eta$ is a learning rate.
Then, puttting these two steps together, we get the update equation
\begin{equation}
    \theta' \leftarrow \theta - \eta g^\mathcal{B} - \sum_{k=1}^C \left( \frac{1}{\lambda_k} - \eta \right) (v_k^\tr g^\mathcal{B})\, v_k.
    \label{eq:learning_update}
\end{equation}
To lower the stochasticity of the estimation of the eigensystem of $\Hlr$ we use an exponential moving average and root mean square for the eigenvalues
\begin{align}
    c_k^{(t)}                     & = \frac{1-\gamma}{1-\gamma^t} \sum_{i=1}^t \gamma^{t-i} c_k^{\mathcal{B}^{(i)}},
                                  &
    c_k^{\mathcal{B}^{(t)}}       & = \frac{1}{|\mathcal{B}_k^{(t)}|}
    \sum_{\mu \in \mathcal{B}_k^{{t}}} \grad z_k^{\mu},                                                                                 \\
    \lambda_k^{(t)}               & = \sqrt{\frac{1-\gamma}{1-\gamma^t} \sum_{i=1}^t \gamma^{t-i} {(\lambda_k^{\mathcal{B}^{(i)}})}^2},
                                  &
    \lambda_k^{\mathcal{B}^{(t)}} & = \frac{1}{|\mathcal{B}^{(t)}|}
    \sum_{\mu \in \mathcal{B}^{(t)}}  y_k^{\mu} \,
    p_k^{\mu} \,(1-p_k^{\mu})  \, |c_k^{\mathcal{B}^{(t)}}|^2,
\end{align}
where $\mathcal{B}_k^{(t)}$ is the set of examples in the $t^{th}$ minibatch, $\mathcal{B}^{(t)}$, that belong to class $k$.
This gives us a hyperparameter, $\gamma$, which needs to take into account how rapidly the high-curvature subspace changes during learning; \citet{GurAri2018} show that the subspace does not markedly change during training, thus, we can set $\gamma$ close to one.
We initialise $c_k^{(0)}=\bm{0}$ but $\lambda_k^{(0)}=1$ since we use $1/\lambda_k$.

\section{Comparison to SGD}
\begin{wrapfigure}[12]{R}{0.5\textwidth}
    \centering
    \includegraphics[width=0.5\textwidth]{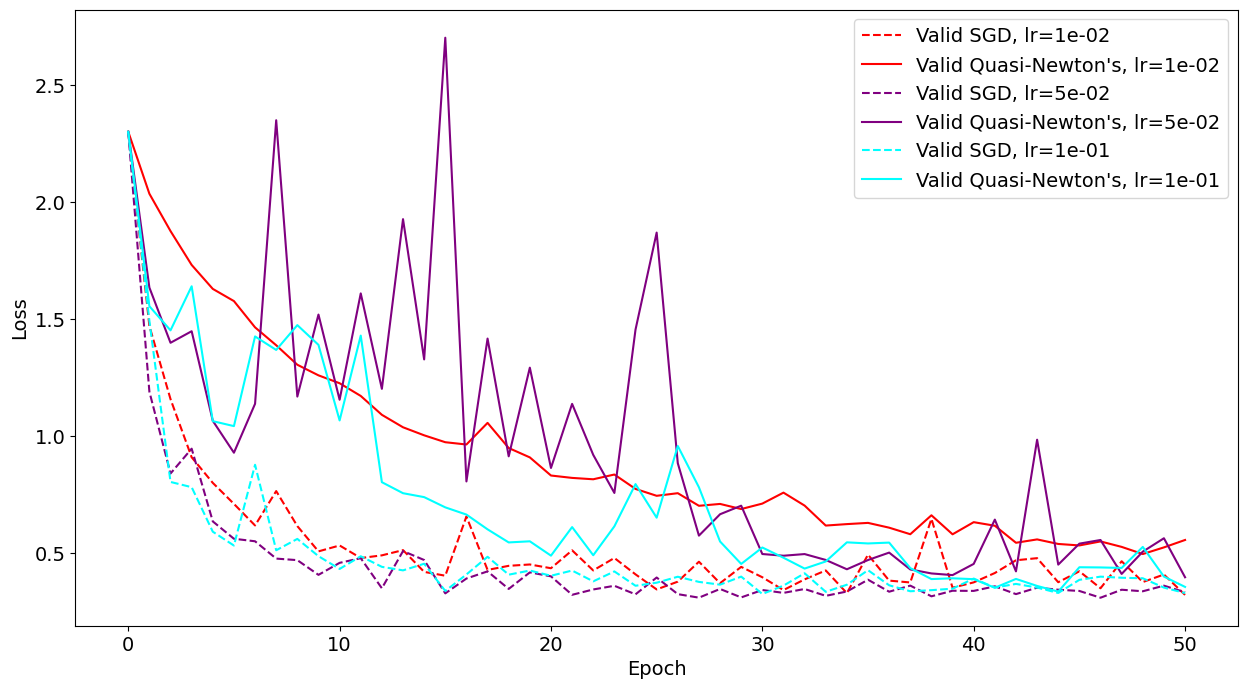}
    \caption{Validation loss for both SGD and quasi-Newton's method}\label{fig:w_wo_hessian_learning}
\end{wrapfigure}

\Cref{fig:w_wo_hessian_learning} shows the typical loss when training a ResNet9 (6.5M parameters) with normal SGD and with the quasi-Newton's method, \cref{eq:learning_update} on CIFAR-10.
We can see that the final performance is similar; however, the quasi-Newton's method is more volatile, slower to improve in general, and does end on average with lower accuracy, see \cref{sec:sgd_qn_comparison}.
One would expect to be able to use a higher learning rate with the quasi-Newton's method as that applies to the low-curvature subspace, however, this is not true, and both methods diverge at $\eta > 10^{-1}$.

\section{Too much noise}
One obvious reason to explore is
If a curve is noisy, then Hessian will capture the noise as local curvature meaning that our estimation will be spurious and oppose learning.

\begin{figure}[ht]
    \centering
    \includegraphics[width=0.75\textwidth]{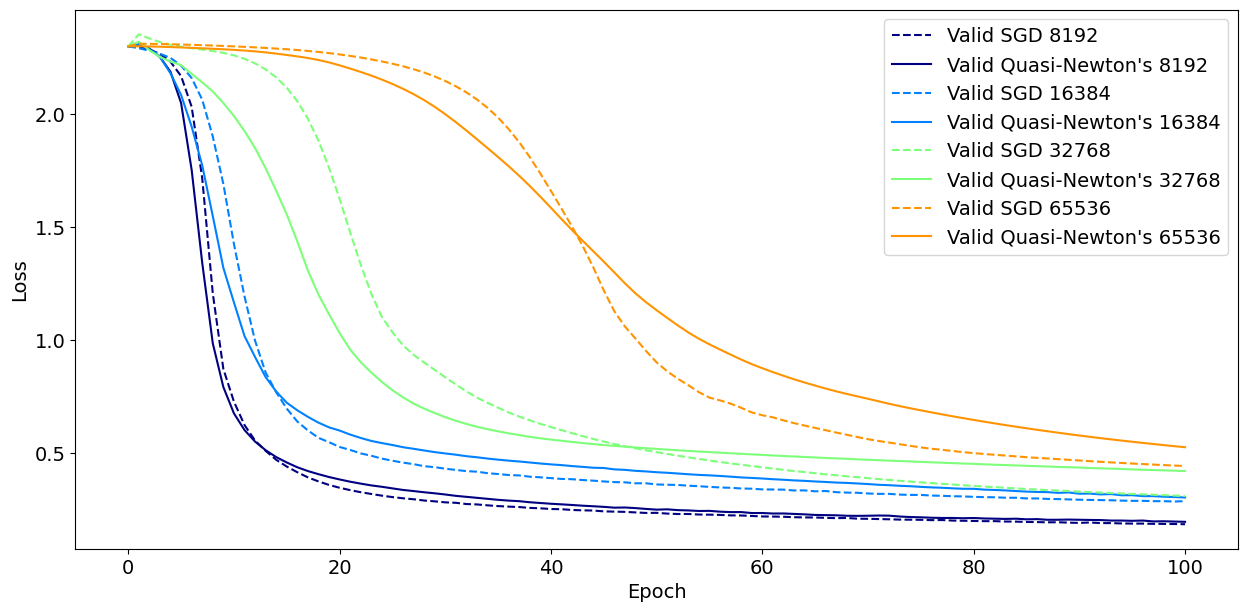}
    \caption{Plots for different batch sizes on MNIST. $\eta = 10^{-2}$}\label{fig:batch_comparison}
\end{figure}

While we have taken steps to reduce noise, \eg{} using the exponential moving average, there might still be too much stochasticity in each minibatch given that the minibatch error is an approximation of the training error which itself is an approximation of the generalisation error.
Indeed, \citet{Granziol2020} shows that ``the extremal eigenvalues of the batch Hessian are larger than those of the empirical Hessian'' due to the noise in the minibatch.

To see if inter-batch noise causes the lack of improvement, we have trained a small CNN on MNIST so that full-batch training is feasible.
From \cref{fig:batch_comparison} we can see that, allthough our quasi-Newton's method may start learning faster than SGD, there comes a crossover point where SGD overtakes and continues to have better performance regardless of the batch size\footnote{Since MNIST only has 50,000 training samples the batch size of 65,535 is just full batch.}.
Smaller batches have their crossover point in the first epoch due to the increased number of steps and so are not shown in \cref{fig:batch_comparison} to illustrate this effect.
Thus, we can conclude that the noise incurred due to batch training is not a root of the performance deficit.

\section{Discussion}
\begin{wrapfigure}[12]{R}{0.5\textwidth}
    \centering
    \includegraphics[width=0.5\textwidth]{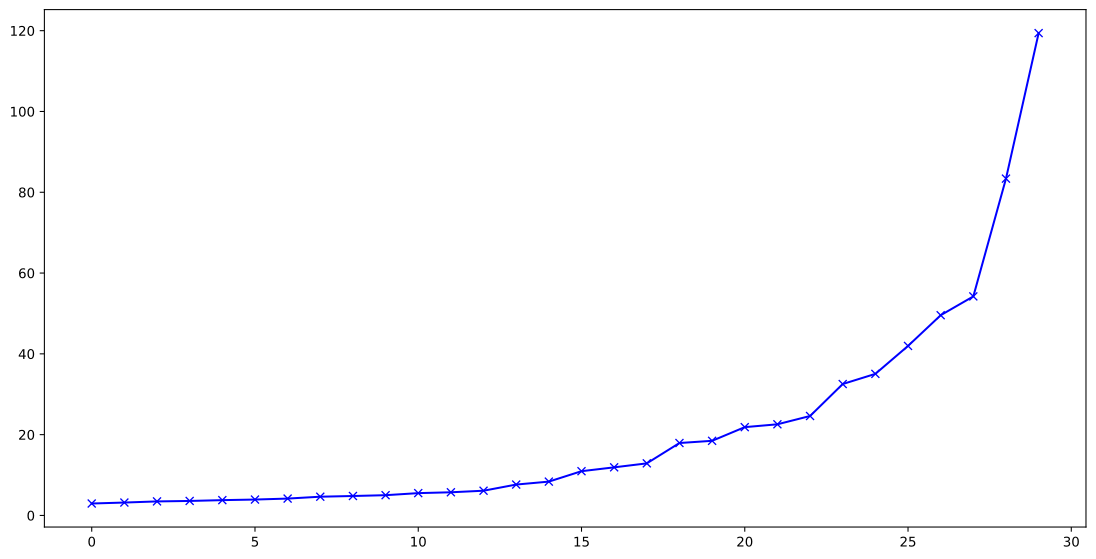}
    \caption{Top 30 true eigenvalues of a small 60k parameter CNN.}\label{fig:top30_eigenvalues}
\end{wrapfigure}

The poor performance of the proposed method is both disappointing and mysterious.  We are currently examining this and it is a work in progress.
We have outlined a few potential explanations for this inconsistency between theory and praxis below.

\subsection{Non-intersection}

If the class vector subspace does not totally cover the top eigenvector subspace, then we are still leaving at least one high-curvature direction not preconditioned. This would also explain the inability to increase the learning rate.

It is interesting to note from \cref{fig:top30_eigenvalues} that, knowing $C$, it seems a reasonable cut-off point for the top-curvature subspace, but if given only the eigenspectrum it would be difficult to guess the number of classes.
This might indicate that $C$ is not the best cut off point for the high-curvature subspace.
It is possible that the class vectors actually have no particular relation to the high-curvature subspace but are simply an arbitrary way of splitting the cumulative gradient into some approximately orthogonal component vectors.
\citet{GurAri2018}, however, argue that it is only the top-$C$ eigenvectors that are preserved over time.
See \Cref{sec:correlation} for more detail on the intersection of these subspaces.

\subsection{Gradient information loss}
It may be that the structure and non-convexity of the landscape is such that rapid gradient descent actually slows down the search.
That is, this method pushes the low-curvature dimensions into a flat region of similar loss, where there is less gradient information, before they can be optimised into a low-loss region, thus slowing down the overall optimisation.

\section{Conclusions}
We are currently investigating the different explanations of why this naive method offers no improvement.
Common experience is that, when done right, optimisation strategies that compensate for different curvatures in the loss landscape can lead to a significant speedup.
Deep learning, however, rules out many traditional methods because of the vast dimensionality of the search space.
Despite the failure of our naive implementation, the recognition that there exists a relatively low-dimensional subspace of high curvature which can be identified relatively easily suggests that it may be possible to improve on current methods.

\bibliography{library}
\clearpage
\appendix
\section{Comparison of SGD and quasi-Newton's}\label{sec:sgd_qn_comparison}
\begin{table}[h!]
  \centering
  \begin{tabular}{r|cc}
     $\eta$ &  SGD & QN \\ \hline
     1e-1 &  88 &  92  \\
     5e-2 &  87 &  92  \\
     1e-2 &  82 &  90  
  \end{tabular}
  \caption{Accuracy (\%)}
\end{table}

\begin{table}[h!]
  \centering
  \begin{tabular}{r|cc}
     $\eta$ &  SGD& QN  \\ \hline
     1e-1 & 0.3574 & 0.3304    \\
     5e-2 & 0.3985 & 0.3346   \\
     1e-2 & 0.5362 & 0.3411
  \end{tabular}
  \caption{Loss}
\end{table}

\section{On the Hessian}

The main disadvantages of most second-order methods are the iteration time complexity, as computing the Hessian is $\mathcal{O}(N^2)$ and then computing its inverse is $\mathcal{O}(N^3)$, and the increased space complexity, $\mathcal{O}(N^2)$, to store the Hessian.
The quasi-Newton's method avoids these pitfalls as it computes an approximation of the inverse Hessian in the same timescale as the gradient, $\mathcal{O}(N)$ and, since our Hessian approximation is formed from outer products, we can calculate a Hessian-vector-product without the quadratic space cost ($H=hh^\tr  \rightarrow Hz = (h^\tr z) h$).
However, it is still infeasible in its current form because it requires $C$ backward passes and storing $c_k$ means we require $C+1$ times as much space as SGD.

Incidentally, it is slightly odd to talk of a Hessian in a landscape with a discontinuous derivative. With ReLU activations we have a continuous piecewise linear surface which has zero Hessians almost everywhere. But even in one dimension, if we approximate a quadratic by a piecewise linear curve, then, although the second derivative is zero almost everywhere, we can
still diverge using gradient descent if the step size is too big since the function is non-locally curved.

\section{Extent of subspace intersection}\label{sec:correlation}

Although there is general consensus that the class vector directions align with the top eigenvectors of the Hessian, we experimentally check to what degree this is true.since this is a probable reason for the worse performance of the quasi-Newton's method.

We collect the gradient directions for each class in the minibatch, $c_k^{\mathcal{B}^{(t)}}$ while training a simple 60k parameter CNN on CIFAR-10.
At the end of training (50 epochs) we calculate the full eigensystem of the trained model.
We can then plot the correlation and check whether the directions are largely stable.

\begin{figure}[ht]
    \centering
    \subfigure[Rank of the combined matrix for each batch]{%
        \includegraphics[width=0.45\textwidth]{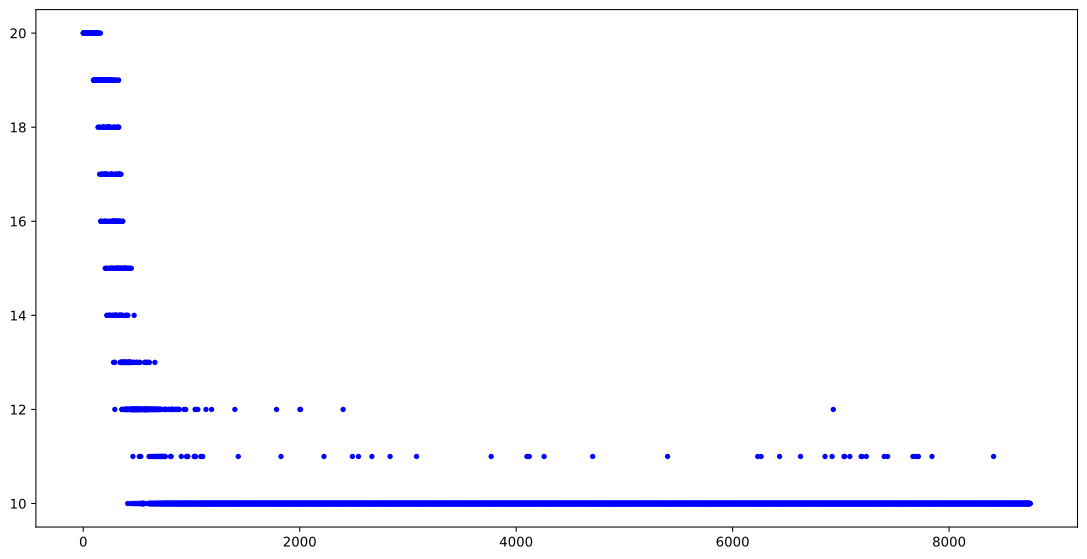}
    }\label{fig:rank_per_batch}%
    \subfigure[Cosine matrix]{%
        \includegraphics[width=0.45\textwidth]{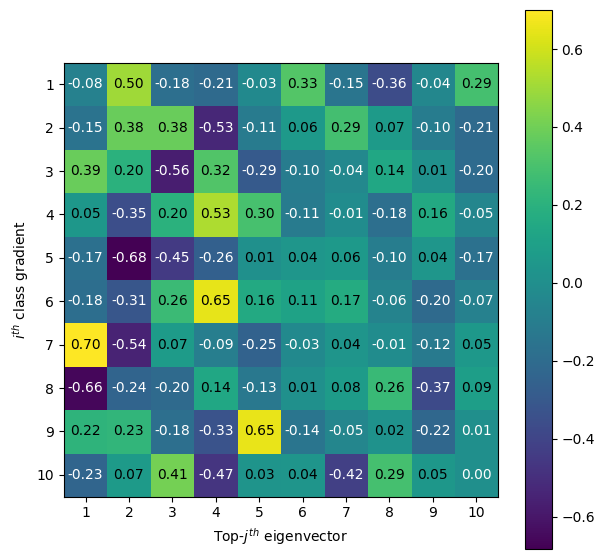}
    }\label{fig:eig_class_corr}
    \caption{Similarity measures between the class gradients and the top-$C$ eigenvectors}
\end{figure}

The cosine-maximised linear assignment for the top-$C$ eigenvectors to the class gradient directions is $\approx 0.36$.
This is a significant correlation since random 60k-dimensional vectors have an expected cosine of order $10^{-3}$.

We can also look at the rank of the combined matrix
\begin{equation}
    [v_1, \ldots, v_C,\; e_0, \ldots ,e_C],
\end{equation}
where $e_i$ is the top $i^\text{th}$ true eigenvector of the fully trained model.
If the subspace spanned by the top-$C$ eigenvectors is the same as that which is spanned by the class gradients then the rank of this matrix should be $C$.
We see that the rank quickly declines from $2C$, \ie no intersection, to consistently being $C$.
However, there are several batches when the rank increases by one or two implying that these batches have a greater amount of noise.

The rank of the combined matrix is a strong indicator that it is covered, but the noise may introduce spurious results due to the precision error in calculating the rank.

\section{Hyperparameters}
For all the experiments we choose a learning rate of 0.1 (unless otherwise specified), basic momentum of with a decay parameter of 0.9, and a gamma 0.9 for the quasi-Newton's method, and 0 weight decay or regularization.
These values were not optimised.

\end{document}